\documentclass{egpubl}
 
\SpecialIssueSubmission    

\CGFccby

\usepackage[T1]{fontenc}
\usepackage{dfadobe}  

\usepackage{cite}  
\BibtexOrBiblatex
\electronicVersion
\PrintedOrElectronic
\ifpdf \usepackage[pdftex]{graphicx} \pdfcompresslevel=9
\else \usepackage[dvips]{graphicx} \fi

\usepackage{egweblnk} 

\usepackage{amsmath}
\usepackage{amssymb}
\usepackage[accsupp]{axessibility}

\usepackage{booktabs} 
\usepackage{caption}
\usepackage{subcaption}
\usepackage{multirow}

\usepackage{amssymb}
\usepackage{pifont}
\usepackage{color}
\definecolor{darkred}{rgb}{0.7,0.1,0.1}
\definecolor{grey}{rgb}{0.65,0.65,0.65}
\definecolor{darkgreen}{rgb}{0.1,0.7,0.1}
\definecolor{cyan}{rgb}{0.7,0.0,0.7}
\definecolor{dblue}{rgb}{0.2,0.2,0.8}
\definecolor{maroon}{rgb}{0.76,.13,.28}
\definecolor{burntorange}{rgb}{0.81,.33,0}
\definecolor{diversity_pink}{rgb}{0.921,.596,0.49}
\definecolor{teaser_grey}{rgb}{0.527,0.527,0.527}
\definecolor{teaser_blue}{rgb}{0.25,0.54,0.71}
\definecolor{teaser_salmon}{rgb}{1.0,0.556,0.491}
\definecolor{iccvblue}{rgb}{0.21,0.49,0.74}
\definecolor{darkred}{rgb}{0.7,0.1,0.1}
\definecolor{darkgreen}{rgb}{0.1,0.7,0.1}
\definecolor{dblue}{rgb}{0.2,0.2,0.8}
\definecolor{maroon}{rgb}{0.76,.13,.28}
\definecolor{burntorange}{rgb}{0.81,.33,0}
\definecolor{cyan}{rgb}{0.0,0.7,0.94}
\definecolor{salmon}{rgb}{0.99,0.51,0.46}
\definecolor{green}{rgb}{0.03,0.91,0.43}
\definecolor{forestgreen}{rgb}{0.13,0.55,0.13}
\definecolor{grey}{rgb}{0.4,0.4,0.4}
\definecolor{purple}{rgb}{0.29,0,0.51}
\definecolor{crimson}{rgb}{0.86,0.08,0.24}

\newcommand*{\ShowNotes}{}

\ifdefined\ShowNotes
  \newcommand{\colornote}[3]{{\color{#1}\textbf{#2: #3}\normalfont}}
\else
  \newcommand{\colornote}[3]{}
\fi

\title[Mixture of Mesh Experts]%
     {Mixture of Mesh Experts with Random Walk Transformer Gating}

\author[Belder \& Tal]
{
\parbox{\textwidth}{\centering
Amir\,Belder\orcid{0000-0002-0729-7252}
and
Ayellet\,Tal\orcid{0000-0002-4967-7309}\\
 Technion - Israel Institute of Technology, Israel
}
}

\begin{document}

\teaser{
\centering
\begin{tabular}{cc}
\includegraphics[width=0.7\textwidth]{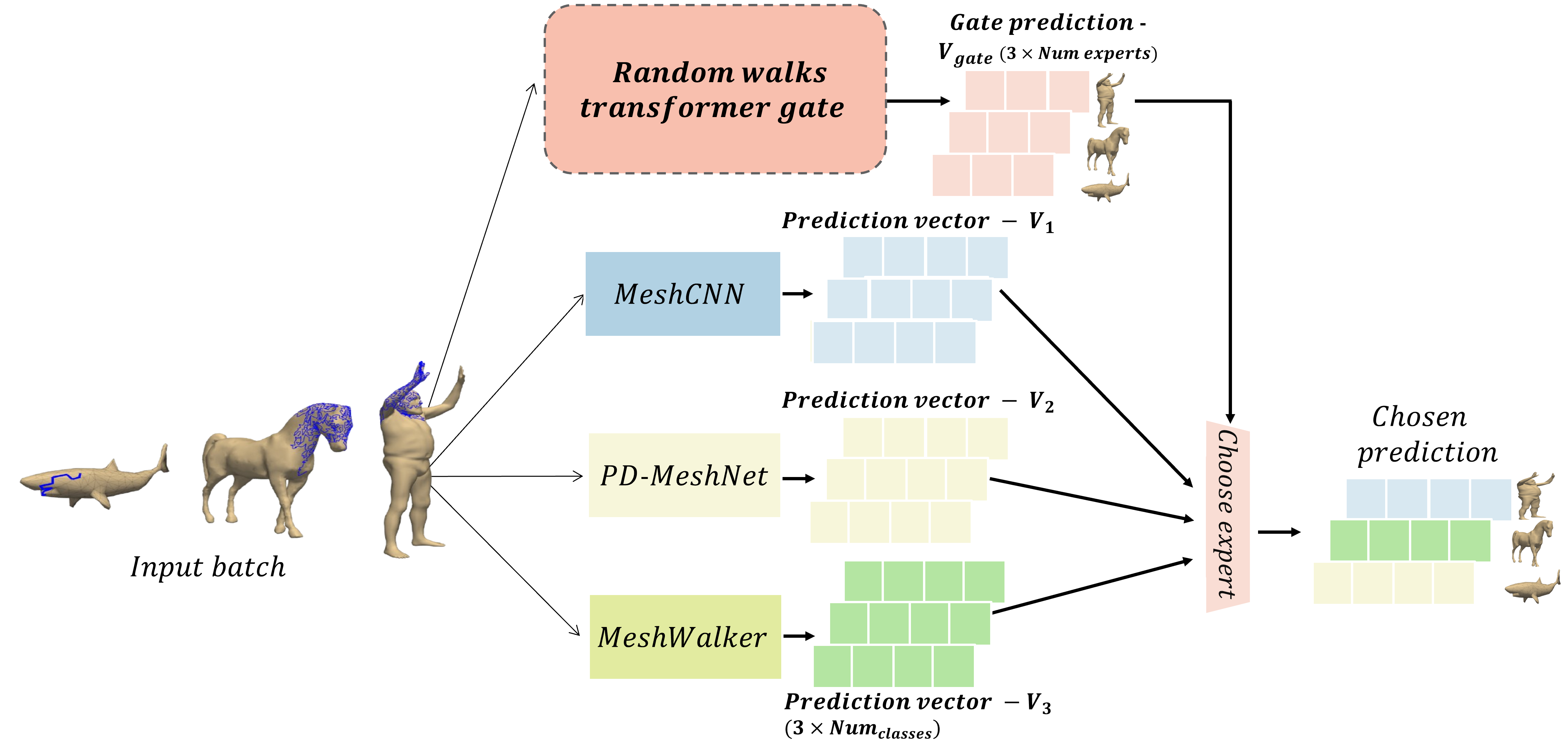} &
\includegraphics[width=0.17\textwidth]{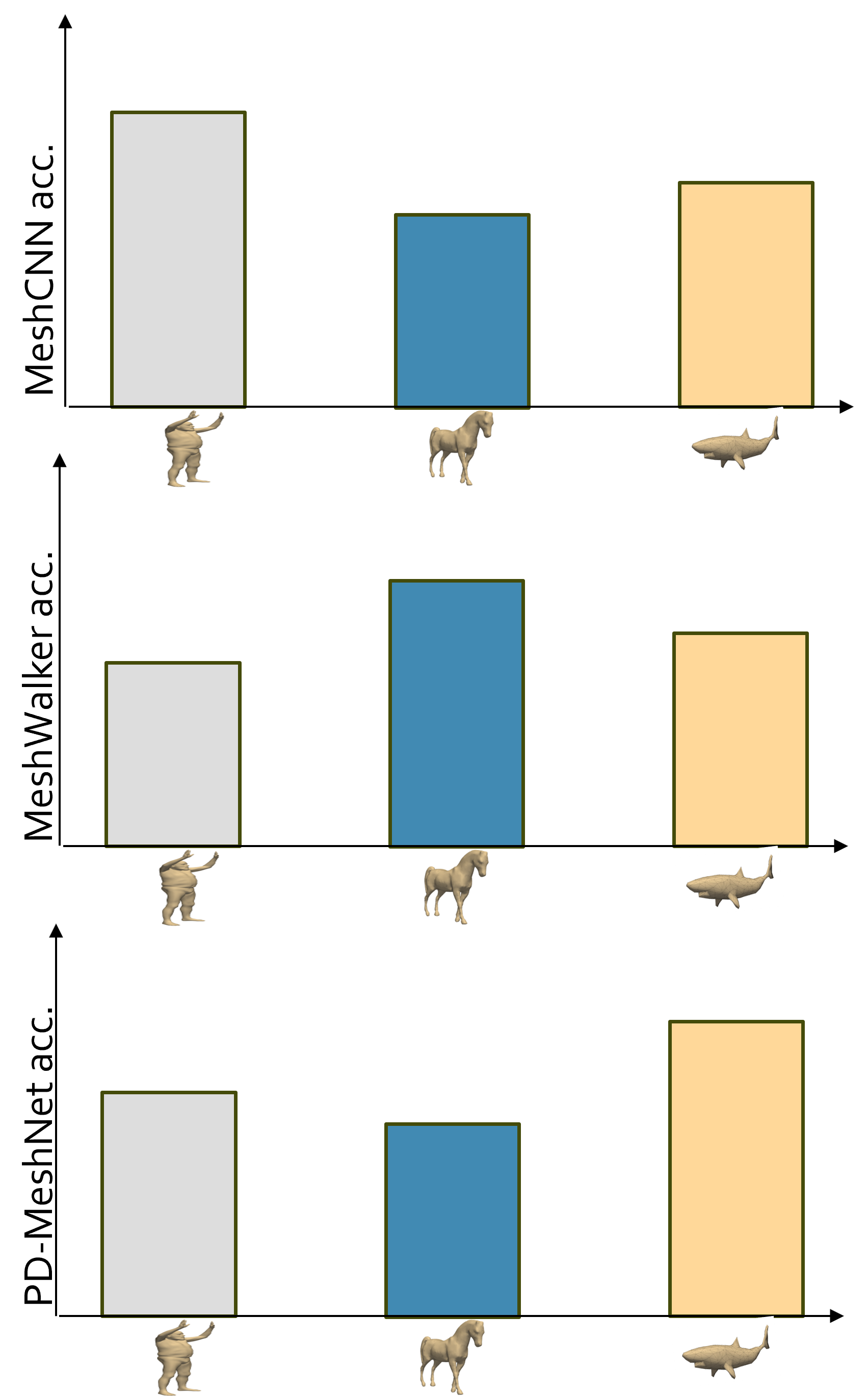} \\
(a)~Mixture of Experts &
(b)~Expert accuracy
\end{tabular}
\caption{\textbf{MoE for mesh analysis.} 
(a) Our Transformer-based random walk gate learns to assign the most suitable expert (e.g., MeshCNN, PD-MeshNet, or MeshWalker) to each input mesh, leveraging the distinct strengths of different architectures. 
By attending to informative regions on the mesh, it identifies areas where specific experts are more accurate, and routes the decision accordingly.
(b) Ideally, the expert that performs best for each class (e.g., Men, Sharks, Horses) should be selected to make the final prediction.
}
\label{fig:teaser}
}

\maketitle
\begin{abstract}
In recent years, various methods have been proposed for mesh analysis, each offering distinct advantages and often excelling on different object classes. 
We present a novel Mixture of Experts (MoE) framework designed to harness the complementary strengths of these diverse approaches. 
We propose a new gate architecture that encourages each expert to specialise in the classes it excels in. 
Our design is guided by two key ideas: (1) random walks over the mesh surface effectively capture the regions that individual experts attend to, and (2) an attention mechanism that enables the gate to focus on the areas most informative for each expert’s decision-making.
To further enhance performance, we introduce a dynamic loss balancing scheme that adjusts a trade-off between diversity and similarity losses throughout the training, where diversity prompts expert specialization, and similarity enables knowledge sharing among the experts.
Our framework achieves state-of-the-art results in mesh classification, retrieval, and semantic segmentation tasks. 
Our code is available at: https://github.com/amirbelder/MME-Mixture-of-Mesh-Experts. 
\end{abstract}



\section{Introduction}
\label{sec:intro}

The polygonal mesh is the most commonly used representation of surfaces in computer graphics, thanks to its numerous advantages, including efficiency and high-quality rendering. 
Given the significance of this representation, numerous methods have been proposed for analyzing meshes, including classification~\cite{mirbauer2021survey}, retrieval~\cite{savva2017large,elad2002content}, and semantic segmentation~\cite{he2021deep,katz2003hierarchical,katz2005mesh}.
Since each method is different, each may excel in different classes.
For example, on SHREC11~\cite{veltkamp2011shrec} ($10/10$ split~\cite{ezuz2017gwcnn}), MeshCNN~\cite{hanocka2019meshcnn} excels in classifying Men, MeshWalker~\cite{lahav2020meshwalker} performs best with Horses, and PD-MeshNet~\cite{milano2020primal} achieves superior classification for Sharks.

In light of this observation, we propose leveraging multiple approaches to maximize results, taking advantage of each model's strengths for the classes it excels in.
There are two common approaches for combining models: {\em Ensembles} and {\em Mixtures of Experts (MoE)}.
Ensemble methods typically aggregate predictions from multiple models using techniques such as averaging~\cite{liu2020ensemble, nam2021diversity} or voting~\cite{dvornik2019diversity, ma2022rethinking, xiang2021walk}.
MoE frameworks consist of multiple expert models, each specialised in handling different parts of an input space, along with a gating network that dynamically assigns input data to the most relevant experts~\cite{cai2024survey}. 
In 3D, MoE has been successfully applied to point cloud normal estimation~\cite{Ben-Shabat_2019_CVPR}, Nerf rendering~\cite{lin2022neurmips,di2024boost}, and point cloud classification using VLMs~\cite{tang2024minigpt, belder2026assistantplacementariabenchmark}.

We introduce a novel method called {\em Mixture of Mesh Experts (MME),} which leverages a Mixture of Experts (MoE) framework for mesh analysis.
MoE systems are defined by their experts and their {\em gate.} 
The gate model learns to make informed predictions by leveraging multiple models.
For instance, in classification, the gate's loss promotes feature {diversity} by encouraging each expert to specialise in specific classes, effectively differentiating their predictions. 
The gate learns which expert performs best for each class, and during inference, it selects the most suitable expert’s prediction for each input example, as shown in Fig.~\ref{fig:teaser}.

We propose a new gate architecture built upon two key insights. 
First, random walks on meshes provide an effective tool for identifying the mesh regions that each expert model focuses on. 
This was previously observed by~\cite{belder2022random}, where these regions were used for adversarial attacks.
Therefore, we employ random walks to our gate architecture.
Second, as each walk may pass multiple regions, we apply attention directly to the random walks, enabling the gate to concentrate on the most relevant regions.

An additional component of our method involves the design of the loss functions. 
Similar to most MoE frameworks, we employ a cross-entropy loss to encourage diversity among experts. 
However, we propose to also use a similarity loss between experts,  allowing experts to learn from one another when it is beneficial.
Encouraging similarity can contradict with the main objective of MoE: deriving each expert to specialise in the classes it performs best on~\cite{xie2024mode}. 
Striking the right balance between these contradictory losses is non-trivial, especially since the true impact on final accuracy is only observable at the end of training.
Our key idea is to dynamically adjust the weighting between diversity and similarity losses throughout training. 
To achieve this, we formulate the problem as a reinforcement learning (RL) task. 
RL is well-suited for this setting, as it is designed to make sequential decisions in continuous environments while optimizing for long-term outcomes.

Our approach has been shown to improve classification, retrieval, and semantic segmentation performance across a range of widely used datasets.
For instance, applying our approach with 
MeshWalker~\cite{lahav2020meshwalker}, MeshCNN~\cite{hanocka2019meshcnn} and PD-MeshNet~\cite{milano2020primal} as experts on SHREC11's~\cite{veltkamp2011shrec} classification attains $100.0\%$ accuracy while each expert alone attains $97.1\%$, $99.1\%$ and $91.0\%$ respectively.
As another example,  when applied with AttWalk~\cite{Ben_Izhak_2022_WACV}, MeshWalker~\cite{lahav2020meshwalker} and MeshNet~\cite{feng2019meshnet} as experts on ShapeNet-Core55's~\cite{savva2017large} retrieval our method improves the overall results by $12.1\%$. 

Hence, our work makes the following contributions:
\begin{enumerate}
    \item 
    We propose a novel method that leverages the strengths of multiple 3D approaches by employing Mixture of Experts (MoE).
    Our method is built upon a novel MoE gate that leverages random walks and applies attention directly to them.
    \item 
   We propose a novel training approach that dynamically balances diversity and similarity using Reinforcement Learning.
    \item 
    Our approach achieves state-of-the-art results for mesh classification, retrieval, and semantic segmentation tasks.
\end{enumerate}

\section{Related work}
\label{sec:related}

\noindent \textbf{Mixture of Experts (MoE).}
This is a learning technique based on the divide-and-conquer principle, where the problem space is partitioned among several neural networks, referred to as experts~\cite{jacobs1991adaptive, nguyen2018practical, richardson2003learning, rokach2010pattern, shazeer2017outrageously, zhou2022mixture}. 
These experts are guided by a gating network that selects which expert to trust for a given input and encourages each expert to develop specialization in certain sub-tasks (e.g, classes), thereby promoting diversity in the feature space.

In Mixture of Experts (MoE) frameworks, the gate often adopts the same architecture as the experts, particularly when the experts are homogeneous~\cite{Dai_2021_CVPR, garbin2021fastnerf, DenseNet2017, He_2016_CVPR, zhong2022metadmoe, NEURIPS2024_b83fae17, lin2022neurmips}. For example, in~\cite{Dai_2021_CVPR}, both the experts and the gate are implemented using ResNet-50~\cite{He_2016_CVPR}. This alignment offers practical advantages, such as ease of implementation, and conceptual ones, including consistent feature processing and representation across the system.
However, when employing heterogeneous experts, i.e., models with differing input modalities or architectural designs, a shared architecture is no longer suitable. 
In such cases, custom gate architectures must be designed to accommodate the variation in expert structures and input representations. 
For instance, in~\cite{pavlitskaya2022evaluating, pavlitska2024towards}, where the experts differ, the gating networks are implemented using shallow 2D convolutional layers. 

The above mentioned works focus mainly on image classification. 
Beyond this scope, MoE has been employed across diverse domains, including natural language processing~\cite{xu2022survey}, 
speech recognition~\cite{negroni2025leveraging,you2021speechmoe}, and multimodal learning~\cite{lin2024moma,li2025uni},   typically utilizing homogeneous expert architectures.
In the 3D domain, MoEs have been investigated for tasks such as normal estimation~\cite{Ben-Shabat_2019_CVPR}, NeRF-based rendering~\cite{lin2022neurmips,di2024boost}, and point cloud classification with vision-language models~\cite{tang2024minigpt}. Notably, these approaches rely on experts that share the same architecture and input modality.
Our work is the first to explore heterogeneous experts for 3D shape analysis, integrating diverse learning models to exploit their complementary strengths and enhance overall performance.

\vspace{0.01in}
\noindent
\textbf{Deep learning for mesh analysis.}
Numerous deep learning models have been recently proposed to address the lack of inherent order and the irregular structure of triangular meshes 
\cite{boscaini2016learning,lei2023s,mirbauer2021survey, verma2018feastnet,wang2022survey}.
They have been used for a variety of tasks including classification~\cite{dai2023mean, feng2019meshnet, fischer2024inemo, gao2025infognn, hanocka2019meshcnn, lahav2020meshwalker, li2022laplacian, milano2020primal}, retrieval~\cite{Ben_Izhak_2022_WACV, feng2019meshnet}, semantic segmentation~\cite{gao2025infognn, hanocka2019meshcnn, lahav2020meshwalker, li2022laplacian, vecchio2023met, Zhong2024MeshSegmenter}, mesh generation~\cite{liumeshdiffusion, nash2020polygen,siddiqui2023meshgpt}, mesh editing~\cite{Decatur_2023_CVPR, gao2023textdeformer}, and adversarial attacks~\cite{belder2022random, fan2023mba, rampini2021universal}.

To demonstrate the generality of our framework, we utilize subsets of six state-of-the-art models as experts, each embodying a distinct design strategy. Specifically:
MeshWalker~\cite{lahav2020meshwalker} leverages random walks across the mesh surface to capture local and global geometric patterns.
MeshCNN~\cite{hanocka2019meshcnn} applies convolutional operations directly on the mesh's edges, preserving its intrinsic structure.
MeshNet~\cite{feng2019meshnet} performs face-based convolutions, emphasizing the geometric properties of mesh faces.
PD-MeshNet~\cite{milano2020primal} integrates attention mechanisms over both faces and edges.
AttWalk~\cite{Ben_Izhak_2022_WACV} combines random walk embeddings using attention to form unified representations.
MeshFormer~\cite{li2022meshformer} treats the mesh as a graph and applies attention mechanisms over its vertices to model complex relationships.
Their combinations highlight the flexibility of our expert-based framework in accommodating diverse architectural paradigms.

\vspace{0.01in}
\noindent
{\textbf{Reinforcement Learning (RL).}} 
RL is used for many applications including robotics~\cite{kober2013reinforcement}, games~\cite{sutton2018reinforcement}, {\em Computer Vision (CV)}~\cite{le2022deep, le2022deep}, {\em Natural Language Processing (NLP)}~\cite{ luketina2019survey,uc2023survey}, and hyper-parameter tuning~\cite{jomaa2019hyp,mazyavkina2021reinforcement}.
RL is a framework for adaptive control, where an agent learns by trial and error to make sequential decisions based on feedback from the environment.
RL problems are defined by their actions, states and rewards that can be either discrete or continuous. 
Each time the agent preforms an action ($a$), the environment responds by preforming a step according to that action and returns an observation (state $s$) and a reward ($r$). 
The actions are selected according to a stochastic policy $\pi$, which determines the probability of choosing an action in the action space.
The state provides information about the environment necessary for the agent to take actions, 
while the reward  encourages the agent to reach convergence.
Value functions~$v$ evaluate the sum of expected future rewards, and are assessed according to a specific policy $v_{\pi}$.

In this work, we adopt the Soft Actor-Critic (SAC) algorithm as our RL framework, not only due to its fast and stable convergence, but also because it is well-suited for environments with continuous state and action spaces, as in our setting~\cite{christodoulou2019soft,haarnoja2018soft}.
SAC extends the standard RL objective of maximizing expected cumulative reward by incorporating an entropy maximization term. 
This encourages the agent to utilize stochastic policies, which has been shown to significantly improve exploration efficiency and overall performance.

\section{Method}
\label{sec:method}

The key ideas of our approach are:
(1) leveraging multiple experts to harness each expert’s strengths for the classes in which it performs best, and
(2) employing two seemingly contradictory loss terms --- similarity and diversity --- and learning how to dynamically balance them during training.

\begin{figure}[t]
\centering
\begin{tabular}{c}
\includegraphics[width=0.47\textwidth]{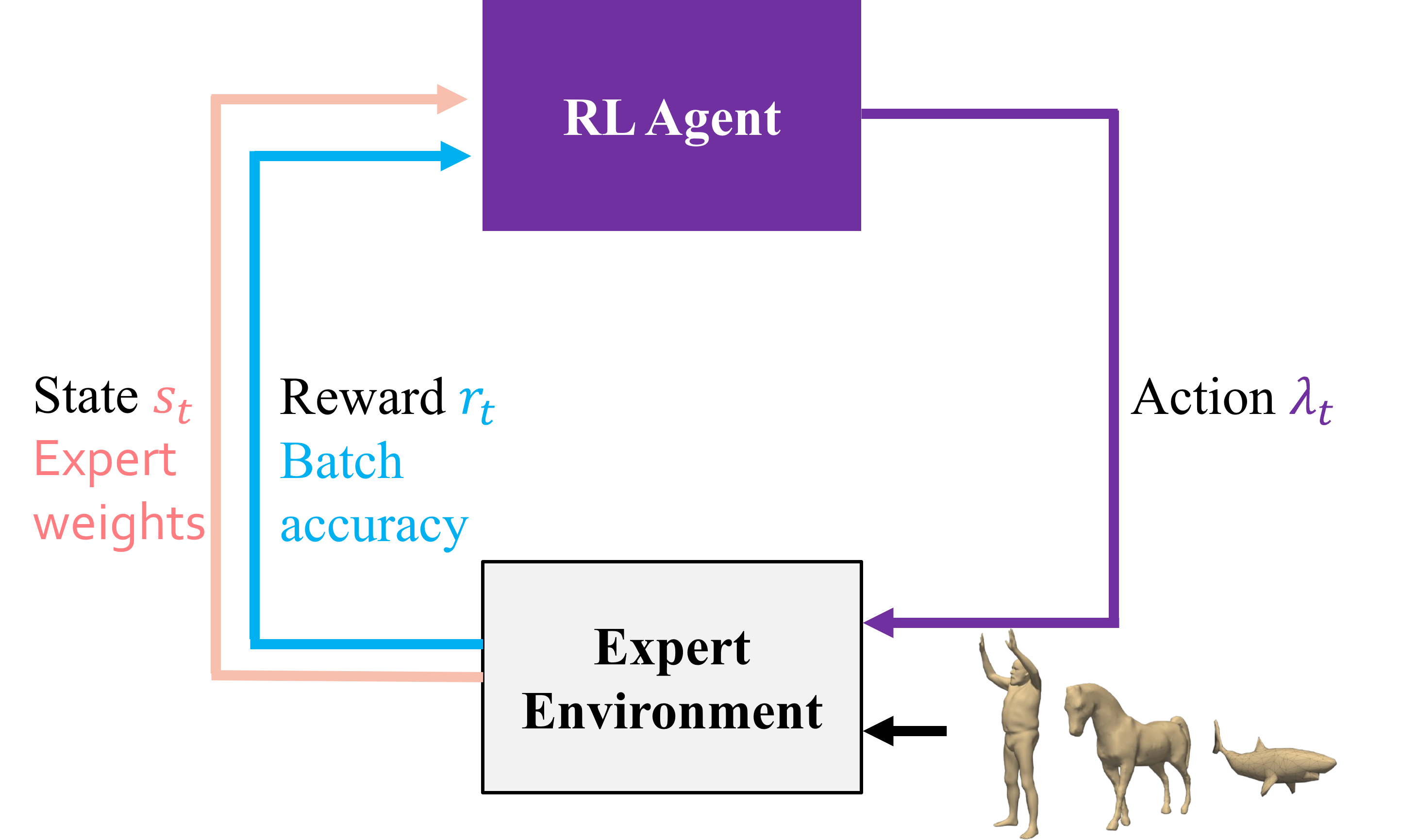}\\
\end{tabular}
\caption{
\textbf{Model.} 
At each iteration ($t$), the expert environment receives a batch of meshes along with a weighting factor ($\lambda_t$) that balances the loss terms. It outputs a weight for each expert ($s_t$) and the batch's accuracy ($r_t$), which are then passed to the agent as input for the next iteration ($t+1$).
}
\label{fig:RL scheme}
\end{figure}

Accordingly, our approach is comprised of two main components: an expert environment and a {\em Reinforcement Learning (RL)} agent, as illustrated in Figure~\ref{fig:RL scheme}.
We formulate the problem in RL terms by defining states, actions, and rewards.
The expert environment is responsible for performing the task, such as classification.
At each iteration $t$ it receives as input a batch of meshes and a weighting factor $\lambda_t$  which balances the two loss terms---similarity and diversity. 
It outputs a weight for each expert $s_t$ (representing the state) and the batch's accuracy $r_t$ (representing the reward). 

The RL agent’s role is to predict the optimal weighting between the loss terms. 
It takes $s_t$ and $r_t$ as input and returns the updated weighting factor $\lambda_{t+1}$ (the action).

\begin{figure*}[htb]
\centering
\begin{tabular}{c}
\includegraphics[width=0.95\textwidth]{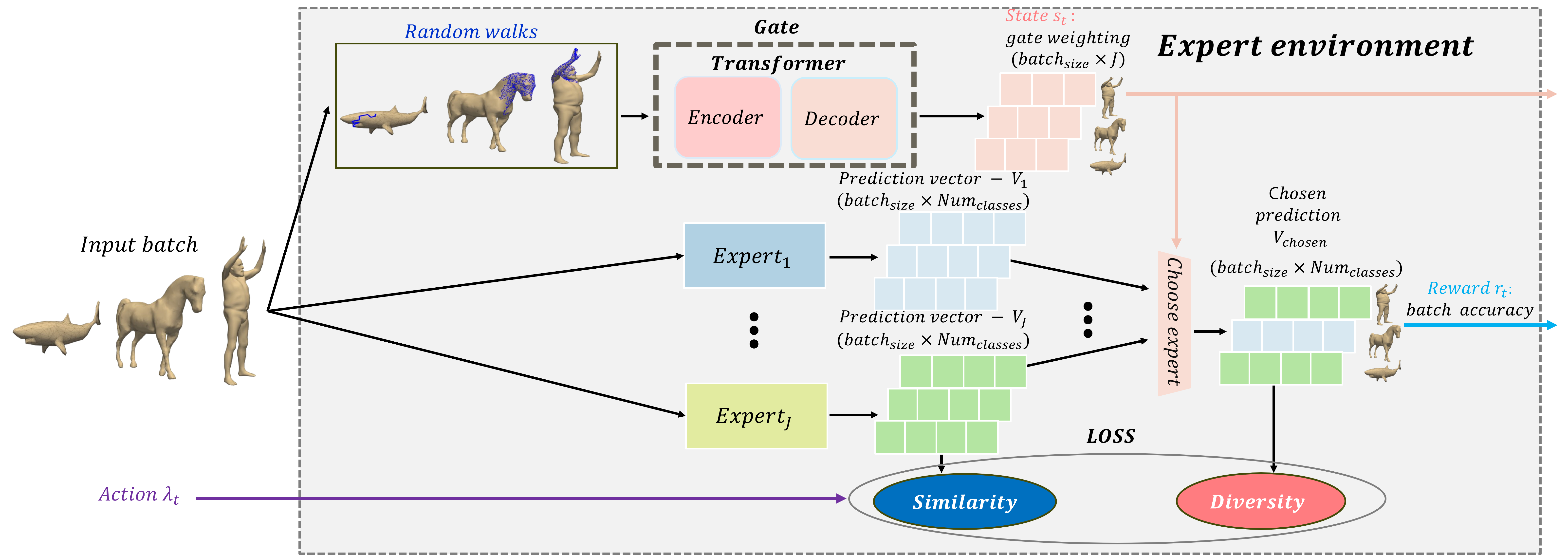}\\
\end{tabular}
\caption{
\textbf{Expert environment.} 
At each iteration $t$, all experts receive the same batch of meshes and independently produce their predictions. 
The same meshes are also fed into the random walk extractor, which generates walk-based representations. 
The extracted random walks are then passed to the Transformer gate as input, which assigns weights to the experts.
The experts' predictions, along with the gate's weights, are passed to the expert chooser, that selects which expert's prediction to use for each mesh, specifically, by choosing the one with the highest assigned weight.
The similarity loss, weighted by $\lambda_t$, is applied to the experts' predictions ($V_1, \dots, V_J$), while the diversity loss is applied to the final prediction, $V_{\text{chosen}}$.
The gate's weightings form the next state ($s_t$) and the batch's accuracy forms the next reward ($r_t$).
}
\label{fig:Full arch}
\end{figure*}

\subsection{Expert environment} 
\label{subsec:environment}

The environment solves the task (e.g., classification) using {\em Mixture of Experts (MoE)}.
As common in MoE setups, it is comprised of a set of experts and a gating mechanism that determines their relative contributions.
The novelty of our model lies primarily in the design of the gating architecture, which is based on two key insights.
First, random walks help to identify the mesh regions that each expert focuses on.
Second, applying attention to these random walks allows the gate to concentrate on the regions most relevant to each expert.
In addition, rather than solely encouraging diversity among experts, as is common in MoE, we also introduce a similarity loss that enables the experts to learn from one another when it is beneficial.

Fig.~\ref{fig:Full arch} provides a detailed view of the expert environment. It takes as input a batch of meshes and the balancing factor $\lambda_t$, which controls the trade-off between loss terms. 
The environment is composed of four main components:
(1) Pre-trained experts, which process the input meshes and output a prediction vector for each mesh.
(2) Random walk extractor, which extracts random walks from each mesh (one example is illustrated in the figure) to be used by the gate.
(3) Gate, implemented as a Transformer, which receives the extracted random walks and assigns a weight to each expert for every mesh.
(4) Expert chooser, which takes as input the experts' predictions and the gate’s weights, and selects the prediction from the expert with the highest weight for each mesh.
The environment outputs the batch's accuracy, denoted as $r_t$, which is computed based on the selected expert predictions ($V_{\text{chosen}}$) and the gate’s weights, $s_t$.
Both the experts and the gate are updated via backpropagation using the two loss terms, weighted by $\lambda_t$.
We elaborate on each component hereafter.

\noindent
\textbf{Experts.}
The environment is composed of $J$ experts, each potentially having a different architecture. 
Leveraging multiple expert models allows us to exploit the strengths of each method on the classes in which it performs best. 
Even when all experts share the same architecture, the MoE mechanism enables them to gain expertise and to specialise in different classes of the dataset, ultimately improving overall performance. 
In our experiments, we employ six expert models: MeshCNN~\cite{hanocka2019meshcnn}, MeshWalker~\cite{lahav2020meshwalker}, PD-MeshNet~\cite{milano2020primal}, AttWalk~\cite{Ben_Izhak_2022_WACV}, MeshFormer~\cite{li2022meshformer}, and MeshNet~\cite{feng2019meshnet}. 
However, our framework is flexible and may utilize other expert architectures as well.

\noindent
\textbf{Random walks.}
Random walks have been shown to be an effective mesh representation for shape analysis~\cite{lahav2020meshwalker}. 
A random walk is defined as a sequence of $L$ distinct vertices, where each consecutive pair is connected by an edge in the mesh. 
We adopt the extraction method of~\cite{lahav2020meshwalker}: given a mesh, the walk begins at a randomly selected vertex. 
Subsequent vertices are added one at a time by randomly selecting a neighbor of the current vertex that has not yet been included in the walk. As a result, each walk consists of $L$ distinct $(x, y, z)$ coordinates corresponding to the vertices it traverses, yielding a representation of size $L \times 3$.

In practice, we set set the sequence length $L$ to $40\%$ of the number of vertices in each mesh.
For training, we employ $8$ random walks per mesh, and during inference, we use $32$ walks per mesh, as in~\cite{lahav2020meshwalker}.
The above hyperparameters are fixed across all datasets and applications.
There is no need to tune them separately for each individual dataset.

\noindent
\textbf{Transformer gate.}
The goal of the gate is to assign a weight to each expert for every input mesh, producing a vector of length $J$ per mesh. 
The key idea is to enable the gate to attend to the mesh regions most relevant to each expert, recognizing that different experts prioritize different areas of the geometry. 
To achieve this, the architecture must first identify these critical regions and then focus on them effectively.
To identify relevant regions, we leverage random walks, as discussed earlier. 
Prior work~\cite{belder2022random} (in the context of adversarial attacks) has shown that random walks serve as strong indicators of important mesh regions and that these regions vary across different networks. 
Moreover, since a single walk can wander around multiple areas on the mesh surface, applying attention along the walk allows the gate to concentrate on the most informative portions of each walk, thereby improving the expert selection.

\begin{figure}[t]
\centering
\begin{tabular}{c}
\includegraphics[width=0.45\textwidth]{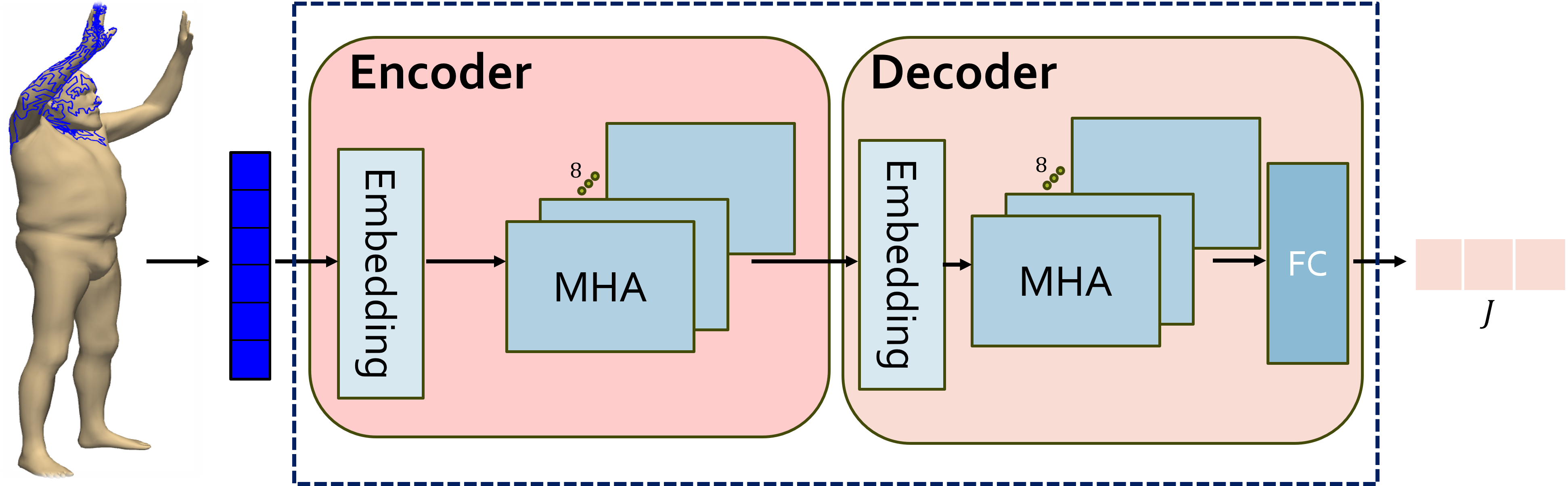}\\
\end{tabular}
\caption{
\textbf{Transformer gate.} We zoom into the gate model.
The gate's input is comprised of a random walk which is extracted from a given input mesh. 
The gate outputs a weight for each of the $J$ experts, which is the key to selecting the prediction of the most suitable expert for the mesh. 
The gate is comprised of an encoder and a decoder.
The encoder processes the walk, and its output serves as input to the decoder.
It consist of a single embedding layer followed by $8$ {\em Multi-Head Attention (MHA)} layers.
The decoder outputs a single weight for each expert (leading to an output at the size of $J$).
It consists of a single embedding layer, followed by $8$ MHA layers, and a FC layer at the size of the number of experts ($J$).
}
\label{fig:model}
\end{figure}

Our gate, depicted in Fig.~\ref{fig:model}, is comprised of an encoder and a decoder. 
The encoder takes a random walk as input and produces an attended representation, where each vertex coordinate is weighted according to its importance. 
It consists of a single embedding layer (a fully connected (FC) layer that projects the input walk into key, query, and value spaces required for attention computation), followed by eight Multi-Head Attention (MHA) layers.
The decoder receives the encoder’s output and generates a weight vector of length $J$, assigning a weight to each expert for the given input mesh. 
It mirrors the encoder's structure: an identical embedding layer, followed by eight MHA layers, and concluding with an FC layer that outputs the final expert weights.
Both the encoder and decoder make use of the scaled dot-product attention mechanism introduced in~\cite{vaswani2017attention}, and all MHA layers follow a shared architectural design throughout the gate.

We note that our gate architecture significantly differs from those used in prior MoE frameworks. 
In settings where all experts share the same architecture, the gate is often designed to mirror the expert architecture itself~\cite{Dai_2021_CVPR, zhong2022metadmoe, NEURIPS2024_b83fae17, lin2022neurmips, Ben-Shabat_2019_CVPR}. 
However, this strategy is impractical when working with heterogeneous expert models.
In image classification tasks involving heterogeneous experts~\cite{pavlitska2024towards, pavlitskaya2022evaluating}, the gate architecture is typically simple, often limited to a single 2D convolutional layer followed by one or two FC layers. 
As we demonstrate in Section~\ref{sec:ablation}, simpler gate designs are less effective in our setting, particularly when expert architectures may vary significantly.

We introduce a pre-training stage aimed at teaching the gate to recognize the mesh regions that each expert relies on for accurate classification. 
This is done through a dedicated training process, applied independently to each expert, during which the gate is trained to "imitate" the expert by producing matching outputs. 
Specifically, given a mesh, a set of random walks is extracted and used as input to the gate, which is trained to predict the expert’s full output vector, that is, the probability distribution over all classes.
Thus, for each expert, the gate gets as input: (1) the meshes of the training set of a dataset (2) the prediction vectors of the expert on these meshes. 
The architecture of the gate, as described earlier, remains unchanged, except that its final FC layer is replaced with one that outputs a vector whose length corresponds to the number of classes in the dataset, rather than the number of experts.
Being able to reproduce the expert's prediction vector has  been shown to reveal the regions the model attends to during classification~\cite{belder2022random}.
This is because, during the imitation process, the regions of interest are more attended and exhibit larger gradients  compared to other parts of the mesh.

To integrate the knowledge learned from all experts, we initialize the final gate by averaging the weights from the expert-specific pre-trained gate models. As will be shown in Section~\ref{sec:ablation}, this initialization strategy contributes to improved overall model accuracy. In practice, the gate is pre-trained separately on each expert for $10$ epochs.

\noindent
\textbf{Expert chooser.}
The purpose of this component is to generate the final predictions for the batch. 
It takes as input the prediction vectors from all experts ($V_1,...,V_J$) along with the corresponding weights produced by the gate ($s_t$). 
For each mesh, it selects the prediction from the expert with the highest weight, resulting in the final output, denoted as $V_{\text{chosen}}$. 
This selection is performed by employing a simple max operation over the gate’s weights.

\noindent
\textbf{Loss.}
An additional aspect of our approach lies in the design of the loss functions. 
Following common practice in MoE frameworks, we employ a diversity loss to encourage specialization among the experts. 
In addition, we introduce a similarity loss that promotes knowledge sharing by allowing experts to learn from one another when beneficial.
To effectively manage the trade-off between these competing objectives, we weigh the similarity loss using a dynamic factor $\lambda_t$, updated at each training iteration. 
This adaptive weighting is crucial, as excessive use of the similarity can hinder the specialization process of the experts~\cite{xie2024mode}.

For the similarity loss, we use the \textit{Kullback--Leibler Divergence (KLD)} to measure the distance between prediction probability vectors. For each mesh and each pair of experts \( j \) and \( w \) (\( w \ne j \)), let \( V_{j} \) and \( V_{w} \) denote the prediction vectors over \( Num_{\text{classes}} \), as computed by \( \text{Expert}_j \) and \( \text{Expert}_w \), respectively.
The similarity loss is computed by applying the KLD between these vectors for every mesh and expert pair:
\begin{equation}
    \mathcal{L}_{\text{sim}} = \frac{1}{B} \sum_{i=1}^{B} \sum_{j=1}^{J} \sum_{\substack{w=1 \\ w \ne j}}^{J} \text{KL}(V_j^{(i)} \,\|\, V_w^{(i)}),
\end{equation}
where \( B \) is the batch size, \( J \) is the number of experts,
\( V_j^{(i)} \) is the prediction vector of expert \( j \) for the \( i \)-th mesh in the batch,
\( \text{KL}(P \,\|\, Q) \) denotes the Kullback--Leibler divergence between distributions \( P \) and \( Q \).

For our diversity loss, we employ the standard MoE gate loss, which is known to encourage diversity among experts~\cite{jacobs1991adaptive}. Specifically, for a batch of size \( B \), let \( V_j^{(i)} \) denote the prediction vector of \( \text{Expert}_j \) for the \( i \)-th mesh in the batch, and let \( s_{t_j}^{(i)} \) be the gate’s weight assigned to \( \text{Expert}_j \) for that mesh. Let \( \mathcal{T}^{(i)} \) denote the ground truth label for the \( i \)-th mesh.
The diversity loss over the batch is defined as:
\begin{equation} 
\mathcal{L}_{\text{div}} = \frac{1}{B} \sum_{i=1}^{B} \sum_{j=1}^{J} s_{t_j}^{(i)} \cdot \text{CrossEntropy}(V_j^{(i)}, \mathcal{T}^{(i)}). 
\label{eq:diversity_batch} 
\end{equation}

Finally, the total loss is defined as a weighted combination of the similarity and diversity losses:
\begin{equation} 
\mathcal{L}_{\text{joint}} = \lambda_t \cdot \mathcal{L}_{\text{sim}} + \mathcal{L}_{\text{div}},
\label{eq:loss}
\end{equation}
where \( \lambda_t \) is the dynamic weighting factor at iteration \( t \).

\subsection{Reinforcement Learning (RL) agent} 
\label{subsec:RLagent}

Instead of relying on a fixed weighting between the diversity and similarity losses, we introduce a learned, adaptive coefficient $\lambda$ that is optimized during training. 
The key challenge is that the influence of $\lambda$ on the model's performance can only be fully evaluated at the end of training, once the final accuracy is known. 
This makes the optimal setting of $\lambda$ a problem that requires considering the entire training process~\cite{belder2023game}.
To address this, we frame the learning process as an RL task, where an agent is trained to predict the appropriate value of $\lambda$ at each iteration. RL is particularly suited to this setting, as it excels in optimizing sequential decisions and handling long-term dependencies~\cite{sutton2018reinforcement, kober2013reinforcement}. 
Its ability to work with delayed reward feedback, such as final accuracy, makes it a natural fit for this problem.
Through experiments, we demonstrate that this approach to employ a {\em dynamic} value of $\lambda$ leads to better results than using a static value throughout the training.

In RL, the agent interacts with an environment by taking actions and receiving feedback in the form of states and rewards. The agent’s goal is to maximize the cumulative expected reward over time. 
This framework is particularly effective in scenarios involving delayed feedback, such as optimizing the final model's accuracy, which is only known after the training is complete.
Thus, we used the accuracy of each batch as the reward. In this way, 
maximizing the expected cumulative reward encourages the agent to discover an effective loss-weighting strategy that ultimately improves the model's accuracy. 
For classification, we used mean instance accuracy; for retrieval, we used mAP; and for segmentation, we used edge accuracy.
It is important to note that both the agent and the dynamic weighting factor $\lambda$ are used solely during training and are not involved at inference time.

In our implementation, we adopt the {\em Soft Actor-Critic (SAC)} RL framework~\cite{haarnoja2018soft, christodoulou2019soft}, selected for its stability and effectiveness in handling continuous state and action spaces which are key requirements in our setting. 
We employ the implementation provided by~\cite{rlalgorithms}.

\section{Results}
\label{sec:results}
In the following, we demonstrate the versatility of our approach by showing that it achieves state-of-the-art results on three fundamental mesh analysis tasks: classification, retrieval, and semantic segmentation.
For each task, we employ three different pre-trained experts. 
Our results are compared against those reported in the original papers.
Furthermore, we compared against a hard (i.e., majority class labels) voting ensemble for each dataset and task, using the expert networks as the ensemble members. 
When available, we selected expert networks based on diverse architectural types (e.g., PD-MeshNet's attention mechanism, MeshCNN's convolutional layers, and MeshWalker's recurrent mechanism) to encourage diversity in their predictions.
To account for the stochastic nature of the walks, all reported results are averaged over three runs.

\subsection{Classification.}
\label{subsec:classification}
Given a mesh, the goal is to classify it into one of several predefined classes.
We evaluate our method on four commonly used datasets for mesh classification: ModelNet40~\cite{wu20153d}, Cube Engraving~\cite{hanocka2019meshcnn}, 3D-FUTURE~\cite{fu20203dfuture}, and SHREC11~\cite{veltkamp2011shrec}.
These datasets vary in object types, number of classes, and the number of meshes per class.
The classification accuracy is defined as the ratio of correctly predicted meshes.

\noindent
\textbf{Datasets.}
{\em SHREC11} consists of 30 classes, each containing $20$ mesh examples.
Following the evaluation protocol of~\cite{ezuz2017gwcnn}, each class is split using either a  $16$/$4$ training/testing configuration or a balanced $10$/$10$ split.
Since most methods achieve $100\%$ accuracy on the $16$/$4$ split, we report results on the more challenging $10$/$10$ split.
{\em Cube Engraving}  dataset contains $4,600$ mesh objects, divided into $3,910$ training samples and $690$ test samples. 
Each object is a cube with a shape “engraved” on one of its faces at a random location.  
The engraved shape is selected from a set of $23$ classes (e.g., turtle, bat, fish), each comprising approximately $20$ unique instances.
{\em ModelNet40} is a widely-used benchmark dataset, comprised of  $12,311$ CAD models across $40$ object categories, with $9,843$ samples used for training and $2,468$ used for testing.
Many objects in ModelNet40 contain multiple disconnected components and are not necessarily watertight, which poses a challenge for some mesh-based methods.
{\em 3D-FUTURE} consists of $9,992$ industrial CAD models of furniture, organized into $7$ super-categories, each containing between $1$ and $12$ sub-categories, for a total of $34$ distinct categories.
The official train/test split includes $6,699$ training models and $3,293$ test models.
This dataset presents a significant challenge due to the intricate geometry of the furniture designs and its hierarchical structure, where visually similar objects often belong to closely related sub-categories which requires fine-grained classification.

\noindent
\textbf{Results.}
For the classification tasks on 3D-FUTURE and ModelNet40, we used three experts: MeshWalker~\cite{lahav2020meshwalker}, AttWalk~\cite{Ben_Izhak_2022_WACV}, and MeshNet~\cite{feng2019meshnet}.
For SHREC11 and Cube Engraving, we employed a different set of three experts: MeshWalker~\cite{lahav2020meshwalker}, MeshCNN~\cite{hanocka2019meshcnn}, and PD-MeshNet~\cite{milano2020primal}.
These expert sets were selected based on their demonstrated effectiveness on the respective datasets and their methodological diversity.
In Section~\ref{sec:ablation}, we further examine how varying the number of experts affects overall performance.

Table~\ref{tbl:classification results 1} presents our results on the 3D-FUTURE and ModelNet40 datasets, alongside the reported performance of other models.
The results for MeshNet and MeshWalker on 3D-FUTURE are taken from~\cite{singh2021meshnet++}.
We observe a significant performance improvement on 3D-FUTURE, demonstrating the effectiveness of our approach on a non-saturated dataset.
On ModelNet40, where many methods already achieve high accuracy, our method yields a more modest gain.
Nonetheless, even on this saturated benchmark, our MoE model outperforms each of its individual experts.
Moreover, although the ensemble outperforms individual experts, it is still less effective than our approach.
Over three runs, the standard deviations of our method are  $0.3\%$  for 3D-FUTURE and $0.1\%$ for ModelNet40.

\begin{table}[tb]
\begin{center}
\small
\begin{tabular}{|l|c|c|}
\hline
Model & 3D-Future & ModlNet40 \\ 
\hline\hline
   MeshNet~\cite{feng2019meshnet} & $64.1\%$ & $91.9\%$\\
MeshWalker~\cite{lahav2020meshwalker}  & $70.2\%$ & $92.3\%$\\
   AttWalk~\cite{Ben_Izhak_2022_WACV}  &  $72.1\%$ & $92.5\%$\\
   MeshNet++~\cite{singh2021meshnet++}  & $71.4\%$ & $91.6\%$  \\
  WalkFormer~\cite{guo2023walkformer} & - & $91.1\%$ \\ 
Ensemble & $78.0\%$ & $92.3\%$ \\
 MME (Ours)  & $\mathbf{86.1\%}$ & $\mathbf{92.9\%}$ \\
\hline
  \end{tabular}
  \caption{
\textbf{Classification results.} 
In this experiment, we used three experts: MeshWalker, AttWalk, and MeshNet.
Our results outperform those of other models and, notably, surpass the individual performance of each expert.
It is also more effective than combining these experts using an ensemble approach.
}
\label{tbl:classification results 1}
\end{center}
\end{table}

Table~\ref{tbl:classification results 2} presents our results on the SHREC11 and Cube Engraving datasets, alongside the reported performance of other models.
Our method achieves $100\%$ accuracy on both datasets. 
This is not surprising, as these benchmarks are considered saturated, with many existing methods already achieving near-perfect results.
Nevertheless, our method still outperforms each of its individual experts, highlighting the benefit of our approach.
Over three runs, the standard deviations are $0.07\%$ for SHREC11 and $0.2\%$ for Cube Engraving.

\begin{table}[tb]
\begin{center}
\small
\begin{tabular}{|l|c|c|}
\hline
Model & SHREC11 & Cube Engraving \\ 

\hline\hline

MeshWalker~\cite{lahav2020meshwalker} & $97.1\%$ & $98.6\%$ \\

AttWalk~\cite{Ben_Izhak_2022_WACV} & $99.7\%$ & - \\

MeshNet++~\cite{singh2021meshnet++} & $99.8\%$ & $98.5\%$ \\

MeshCNN~\cite{hanocka2019meshcnn} & $91.0\%$ & $92.2\%$ \\

GWCNN~\cite{ezuz2017gwcnn} & $90.3\%$ & - \\

PD-MeshNet~\cite{milano2020primal} & $99.1\%$ & $94.4\%$ \\

MEAN~\cite{dai2023mean} & $99.1\%$ & $98.9\%$ \\

WalkFormer~\cite{guo2023walkformer} & $98.8\%$ & $99.4\%$ \\
  
Ensemble & $99.9\%$ & $99.4\%$ \\

 MME (Ours)  & $\mathbf{100.0\%}$ & $\mathbf{100.0\%}$ \\
\hline
  \end{tabular}
  \caption{
\textbf{Classification results.}
In this experiment, we used three experts: MeshWalker, MeshCNN, and PD-MeshNet.
Our results not only outperform those of other models but also exceed the individual performance of each expert and the ensemble.
}
\label{tbl:classification results 2}
\end{center}
\end{table}

\subsection{Retrieval.}
\label{subsec:retrieval}
Given a query mesh, the objective is to retrieve and order the dataset objects based on their similarity to the query. 
ShapeNet-Core55~\cite{savva2017large} \& ModelNet40~\cite{wu20153d} are the commonly used datasets for this task.
The primary evaluation metric is the {\em mean average precision (mAP)}, mainly used for ModelNet40. 
For ShapeNet-Core55, other measures are also employed, most notably the {\em Normalized Discounted Cumulative Gain (NDCG)}.

\noindent
\textbf{Datasets.}
{\em ShapeNet-Core55} is a subset of ShapeNet~\cite{savva2016shrec16}, containing  $51,162$ meshes across $55$ categories, where   $35,764$  samples are used for training and $10,265$ are used for testing.
We also use {\em ModelNet40,} as described above.
Following the protocol in~\cite{savva2017large}, we retrieve up to $1,000$ meshes per query.

\noindent
\textbf{Results.}
For retrieval on ShapeNet-Core55 and ModelNet40, we used three experts: MeshWalker~\cite{lahav2020meshwalker}, AttWalk~\cite{Ben_Izhak_2022_WACV}, and MeshNet~\cite{feng2019meshnet}.
These models were selected due to their strong performance on these datasets and the methodological diversity they represent.

Table~\ref{tbl:retrieval} presents our retrieval results alongside the reported performance of other models.
Since MeshWalker did not report results for this task, we include results from our own reproduction.
On ShapeNet-Core55, our approach achieves a substantial improvement over existing methods.
On ModelNet40, on which many methods already attain high accuracy, our method yields a smaller but consistent gain, as discussed earlier.
Over three runs, the standard deviations are $0.1\%$ for both ShapeNet-Core55 and ModelNet40.

\begin{table}[t]
\begin{center}
\small
\begin{tabular}{|l|c|c|c|}
\hline
 Method & \multicolumn{2}{c|}{ShapeNet-Core55} & \shortstack{ModelNet40} \\

\cline{2-4}
&   \scriptsize mAP &   \scriptsize NDCG &   \scriptsize mAP  \\
\hline\hline

{\footnotesize AttWalk~\cite{Ben_Izhak_2022_WACV}} &    $81.1\%$ &    $86.7\%$ &    $91.2\%$  \\

{\footnotesize MeshWalker~\cite{Ben_Izhak_2022_WACV}} &    $79.2\%$ &    $84.5\%$ &    $89.9\%$  \\

{\footnotesize DALN~\cite{savva2017large}}  &    $66.3\%$ &    $76.2\%$ & - \\

 {\footnotesize  MeshNet~\cite{feng2019meshnet}} &  $70.2\%$ &     $75.3\%$  &      $81.9\%$ \\
     
{\footnotesize Ensemble}&    $84.3\%$ &    $85.6\%$ &    ${89.2\%}$ \\

 {\footnotesize MME (Ours) } &   $\mathbf{93.2\%}$ &    $\mathbf{93.8\%}$ &    $\mathbf{92.9\%}$ \\
   \hline 
\end{tabular}
\caption{
\textbf{Retrieval results.}
In this setting, we use MeshWalker, AttWalk, and MeshNet as expert models.
We note that the results of MeshNet on ShapeNet-Core55 are from our own reproduction, as they are not reported in the original paper.
Our results outperform those of other models and surpass the individual performance of each expert, as well as the ensemble.
}
\label{tbl:retrieval}
\end{center}
\end{table}

\subsection{Semantic segmentation.}
\label{sub:segmentation}
The goal of mesh semantic segmentation is to determine, for the basic elements of the mesh, to which segment they belong.
We use widely adopted benchmark datasets for evaluation:  Human Body~\cite{maron2017convolutional},
 COSEG~\cite{wang2012active},  and ++PartNet~\cite{mo2019partnet}.
The segmentation is assessed based on the percentage of correctly labeled faces or edges  (depending on the dataset) assigned to segments, as follows.
Face accuracy is defined as the percentage of correctly-labeled mesh faces~\cite{Haim_2019_ICCV}.
Edge accuracy is defined as the accuracy of all edges, weighted by the edge’s length, using the edge-level semantic labeling of~\cite{hanocka2019meshcnn}.

\noindent
\textbf{Datasets.}
The {\em Human body} dataset consists of $370$ training models and $18$ test models.
The meshes are manually segmented into eight labeled segments.
{\em COSEG} consists of three large classes: aliens, vases, and chairs, containing $200$, $300$, and $400$ shapes, respectively.
Each category is split into $85\%$ training and $15\%$ testing sets. 
{\em PartNet} consists of $26{,}671$ models across $24$ categories.
Each category is split into $70\%$ training, $10\%$ validation, and $20\%$ testing sets.

\noindent
\textbf{Results.}
For semantic segmentation, we use three expert models: PD-MeshNet~\cite{milano2020primal}, MeshFormer~\cite{li2022meshformer}, and MeshCNN~\cite{hanocka2019meshcnn}.

Table~\ref{tbl:segmentation res} shows our results, alongside the reported performance of other models. 
(The results on PartNet are from our own reproduction, as they were not provided.)
On the Human Body dataset, our method achieves an improvement of nearly $2\%$.
On COSEG, which is considered saturated, the improvement is modest.
On PartNet, the least saturated dataset, our method achieves an improvement of $6.7\%$.
As before, our approach outperforms each of its individual experts.
Over three runs, the standard deviations are $0.3\%$ for COSEG,  $0.3\%$ for Human Body, and 0.2\% for PartNet.

\begin{table}[t]
\begin{center}
\small
\begin{tabular}{|l|c|c|c|c|c|}
\hline
 \multirow{2}{*}{Model} & \multicolumn{2}{c|}{Human body} & \shortstack{COSEG} & \multicolumn{2}{c|}{PartNet} \\
\cline{2-6}
& \scriptsize Face & \scriptsize Edge & \scriptsize Edge & \scriptsize Face & \scriptsize Edge \\
 \hline
 \hline
 MeshWalker & $92.7\%$ & $94.8\%$ & $99.1\%$ & - & - \\
 WalkFormer & - & $98.0\%$ & $99.4\%$ & - & - \\
 MeshCNN & $89.0\%$ & $92.3\%$ & $98.2\%$ & $61.0\%$ & $62.1\%$ \\ 
 MeT & - & $93.6\%$ & $99.3\%$ & - & - \\
 PD-MeshNet & $85.6\%$ & - & $96.9\%$ & $58.7\%$ & $59.8\%$ \\
 MeshFormer & - & - & $99.0\%$ & $64.0\%$ & $64.3\%$ \\
 DiffusionNet & - & $95.5\%$ & - & - & - \\
 Ensemble & $92.0\%$ & $98.5\%$ & $99.0\%$ & $64.5\%$ & $66.2\%$ \\
 MME (Ours) & $\mathbf{94.5\%}$ & $\mathbf{99.7\%}$ & $\mathbf{99.9\%}$ & $\mathbf{69.9\%}$ & $\mathbf{71.0\%}$ \\
\hline
\end{tabular}
\caption{
\textbf{Semantic segmentation accuracy.}
In this experiment, we used three experts: MeshFormer, MeshCNN, and PD-MeshNet.
Our results outperform those of other models and surpass the individual performance of each expert as well as the ensemble.
We note that since the competing methods do not report results on PartNet, we ran their publicly available code to obtain the results used as our expert baselines.
}
\label{tbl:segmentation res}
\end{center}
\end{table}

\begin{figure}[t]
\centering
\begin{tabular}{c}
\includegraphics[width=0.28\textwidth]{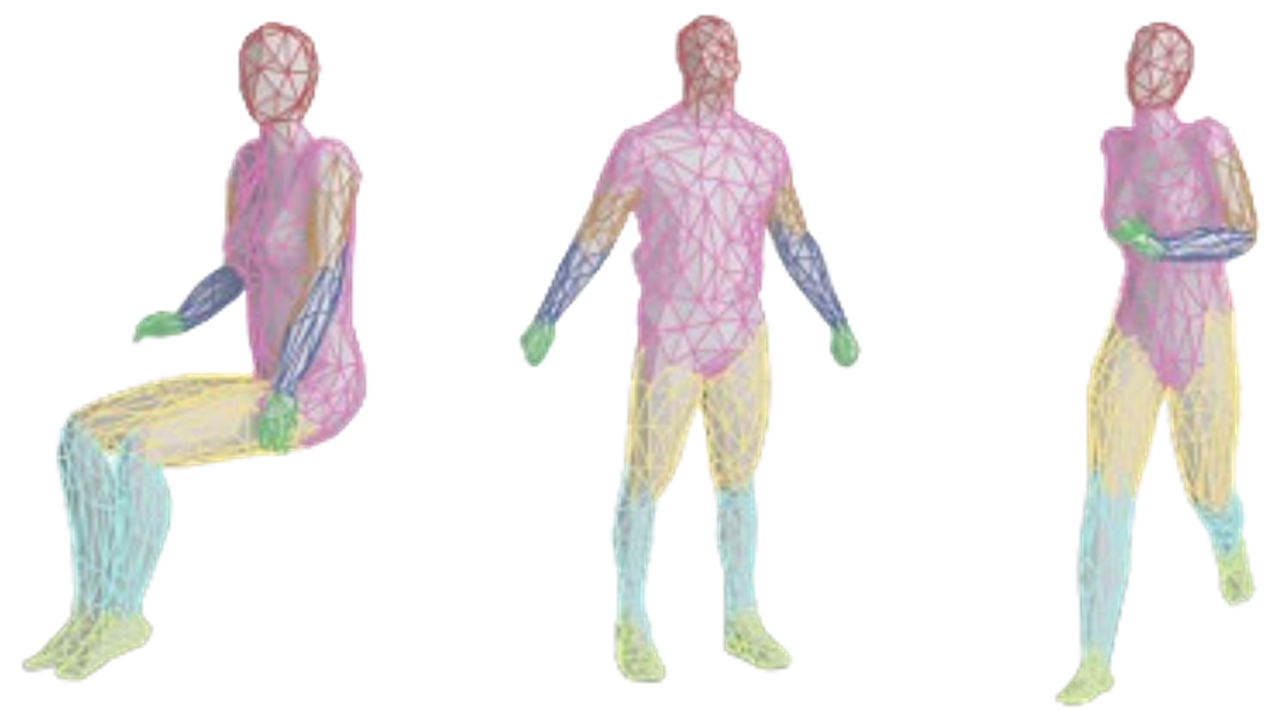}
\end{tabular}
\caption{
\textbf{Qualitative results.} Body parts are clearly segmented and consistently labeled across poses.
}
\label{fig:human}
\end{figure}

Fig.~\ref{fig:human} provides qualitative examples on the Human body dataset.
The different body parts are accurately distinguished and segmented in each mesh, with consistent labeling across varying poses.

\begin{figure}[t]
\centering
\begin{tabular}{ccc}
\includegraphics[width=0.1\textwidth]{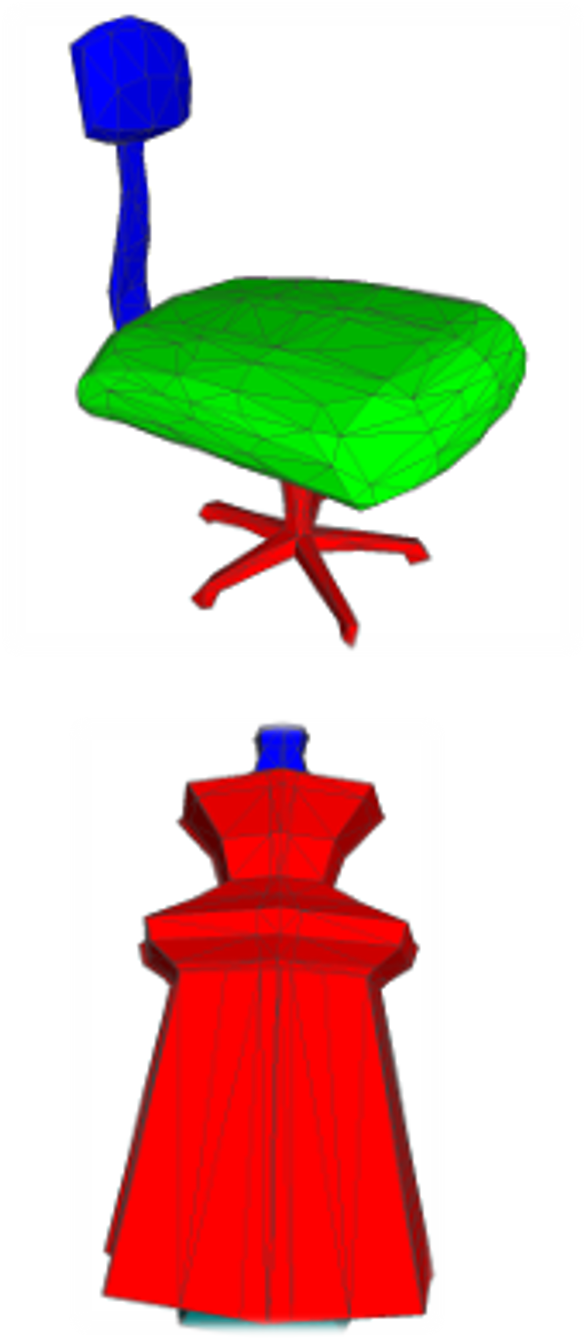} & \includegraphics[width=0.1\textwidth]{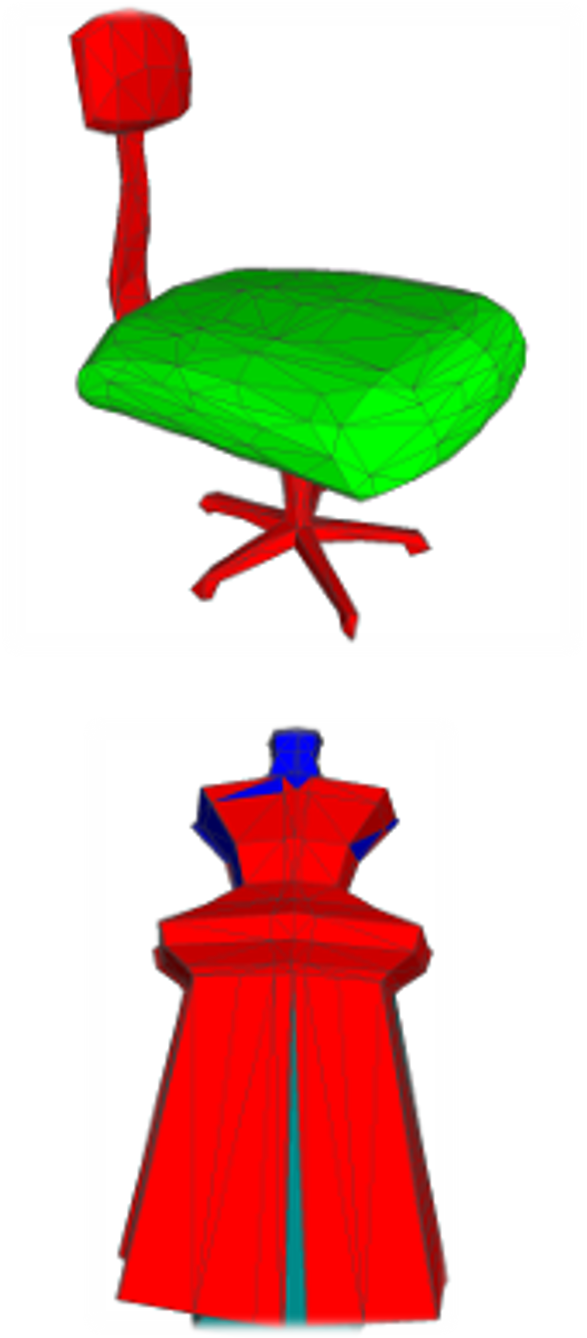} & \includegraphics[width=0.1\textwidth]{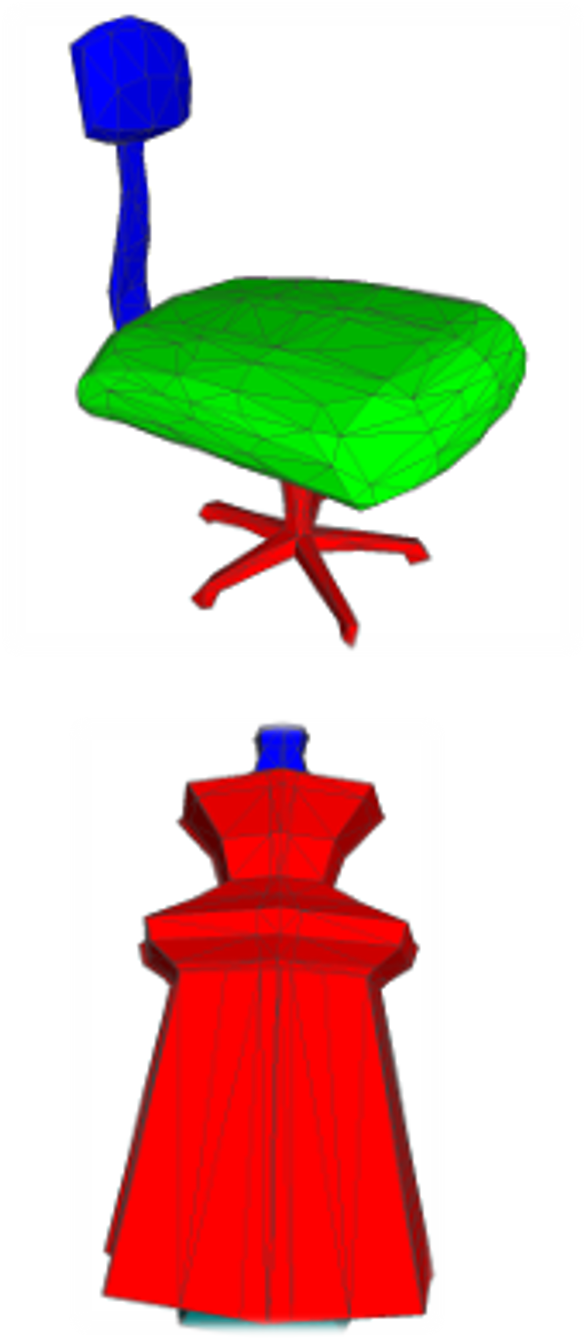} \\
(a) Ground truth & (b) PD-MeshNet & (c) Ours 
\end{tabular}
\caption{
\textbf{Qualitative results of segmentation (COSEG).} 
Our segmentations closely match the ground truth, whereas PD-MeshNet produces errors on these objects.
This is because our model selected MeshCNN as the expert for these cases, rather than PD-MeshNet.
}
\label{fig:coseg}
\end{figure}

Fig.~\ref{fig:coseg} shows examples where our method corrects segmentation errors made by one of the experts, PD-MeshNet.
Specifically, in the chair example, a leg is mistakenly classified as part of the backrest, and in the vase example, the inner region is misclassified, and one triangle at the top incorrectly labeled as the neck.
Our method aligns better with the ground truth because the network selected a different expert (MeshCNN) for these particular inputs.

\section{Ablation Study}
\label{sec:ablation}
To gain clearer insight into the contributions of individual components within our framework, we performed a series of ablation studies. In particular, we examined the impact of optional gating mechanisms, parameter choices, and expert considerations, and we also report the running times and limitations. All experiments were carried out on the 3D-FUTURE dataset, which remains unsaturated in terms of performance, with some experiments conducted also on other datasets.

\noindent
\textbf{On the choice of the gate.}
Our goal is to evaluate whether our proposed gating mechanism, based on attention applied directly to random walks, is more effective than alternative designs.
To this end, we replaced our gate with the following classification networks, where the classification task is to select among the $J$ experts:
(1)~A gate composed of two FC layers.
(2)~A gate consisting of a 3D convolutional layer followed by two FC layers, inspired by gating mechanisms used in image classification~\cite{pavlitska2024towards, pavlitskaya2022evaluating}.
(3)~MeshNet~\cite{feng2019meshnet}, which performs convolutions on mesh faces.
(4)~MeshWalker~\cite{lahav2020meshwalker}, which operates on random walks but does not use attention.
(5)~PD-MeshNet~\cite{milano2020primal}, which applies attention mechanisms to mesh faces and edges instead of walks.
(6)~AttWalk~\cite{Ben_Izhak_2022_WACV}, which uses attention to aggregate MeshWalker embeddings into a unified representation.
{When using a network as a gate, we preserve its original structure and modify only the final FC layer so that its output dimension corresponds to the number of experts rather than the number of classes.
}

Table~\ref{tbl:different gates} shows that, as expected, the simplest gate designs result in the lowest performance, whereas more sophisticated classification networks achieve significantly better results.
These results also suggest that random walk-based methods are particularly effective for this task.
Our approach outperforms that of AttWalk~\cite{Ben_Izhak_2022_WACV}, whose objective is to generate a single global representation from all walk embeddings.
In contrast, our goal is to focus on all informative regions in order to guide the selection of the most appropriate expert.
We note that AttWalk and our method (MME) are comparable in size, with $12{,}703{,}011$ and $12{,}682{,}002$ trainable parameters, respectively.

\begin{table}[tb]
\begin{center}
\begin{tabular}{|l|c|c|c|}
\hline
Gate & Classi. & Seg. & Retrieval \\ 
\hline\hline
$2$ FC  & $74.2\%$ & $92.0\%$ & $83.4\%$ \\
3D conv. + $2$ FC  & $76.4\%$ & $92.2\%$ & $84.5\%$ \\
MeshNet~\cite{feng2019meshnet} & $80.0\%$ & $92.7\%$ & $85.1\%$ \\
PD-MeshNet~\cite{milano2020primal} & $81.7\%$ & $93.5\%$ & $85.5\%$ \\
MeshWalker~\cite{lahav2020meshwalker} & $83.8\%$ & $93.8\%$ & $86.9\%$ \\
AttWalk~\cite{Ben_Izhak_2022_WACV} & $84.6\%$ & $94.0\%$ & $90.5\%$ \\
MME (Ours) & $\mathbf{86.1}\%$ & $\mathbf{94.5}\%$ & $\mathbf{93.2}\%$ \\
\hline
\end{tabular}
\caption{
\textbf{Different gates.}
Our method’s gating mechanism achieves the best results compared to other gating alternatives.
In this experiment, we evaluated it on 3D-FUTURE classification, Human Body segmentation, and ShapeNet-Core55 retrieval.
}
\label{tbl:different gates}
\end{center}
\end{table}

\begin{figure}[t]
\centering
\begin{tabular}{c}
\includegraphics[width=0.3\textwidth]{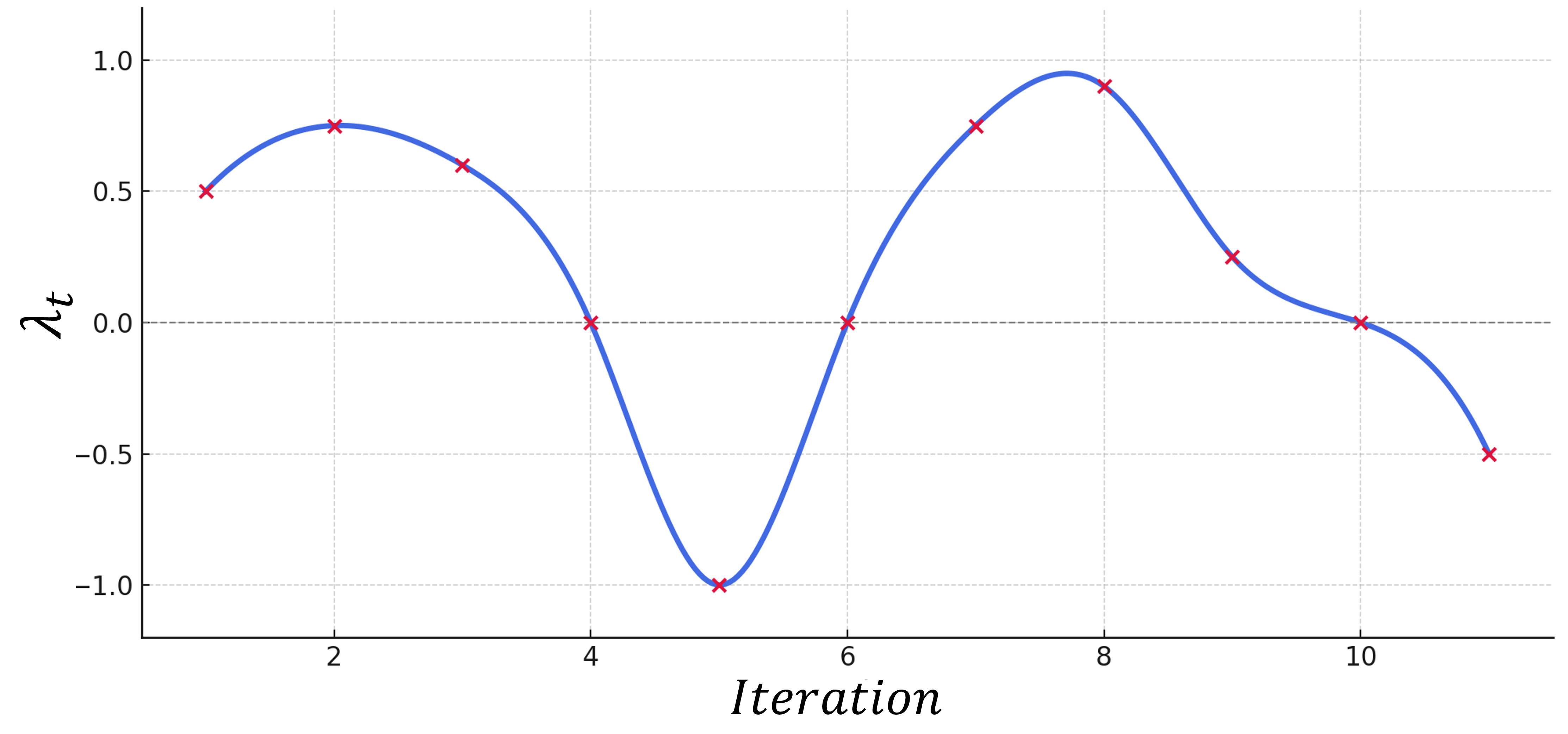} \\ 
\end{tabular}
\caption{
\textbf{$\lambda$'s dynamic value}. 
The value of $\lambda$ changes dynamically along 3D-FUTURE's training, transitioning between positive values that promote similarity and negative values that encourage diversity, in order to achieve optimal accuracy.
}
\label{fig:lambda value}
\end{figure}

\noindent
\textbf{On the dynamic $\lambda$.}
We evaluate whether a dynamic $\lambda$ outperforms a fixed one.
Fig.~\ref{fig:lambda value} illustrates that $\lambda$ varies throughout training, transitioning between positive values that encourage similarity among experts and negative values that promote diversity.
Table~\ref{tbl:Importance of a dynamic lambda value} confirms that our dynamic $\lambda$ strategy outperforms all constant $\lambda$ settings, including $\lambda = 0$, which corresponds to the vanilla MoE that encourages only diversity among experts.

\begin{table}[t]
\begin{center}
\begin{tabular}{|l|c|}
\hline
$\lambda$'s Value & Accuracy \\ 
\hline
\hline
$-1$ & $70.1\%$\\ 
$0$ & $74.8\%$ \\ 
$0.01$ & $82.3\%$ \\ 
$0.1$ & $76.3\%$ \\ 
$0.5$ & $76.7\%$ \\  
$1$ & $77.1\%$ \\ 
Dynamic (Ours) & $\mathbf{86.1\%}$ \\ 
\hline
\end{tabular}
\caption{
\textbf{Comparison with static $\lambda$ values.}
Our dynamic $\lambda$ strategy achieves the highest accuracy
}
\label{tbl:Importance of a dynamic lambda value}
\end{center}
\end{table}

\begin{table}[t]
\begin{center}
\begin{tabular}{|l|c|}
\hline
Loss & Accuracy \\ 
\hline
\hline
Diversity & $82.3\%$ \\
Diversity + Cosine & $84.0\%$ \\ 
Diversity + MSE & $82.9\%$ \\
Diversity + KLD (Ours) & $\mathbf{86.1\%}$ \\
 \hline
\end{tabular}
\caption{
\textbf{Accuracy with different losses.}
KLD as the similarity loss achieves the highest accuracy, outperforming alternative similarity losses and diversity-only training.
}
\label{tbl:different similiarity losses}
\end{center}
\end{table}

\begin{table}[t]
\begin{center}
\small
\begin{tabular}{|c|c|c|c|l|}
\hline
 \multirow{2}{*} {Class} & \multicolumn{3}{c|}{Accuracy} & \multirow{2}{*} {Expert choice} \\ 
 \cline{2-4}
& AttWalk & MeshWalker & MeshNet & \\
\hline\hline
Armchair & $\mathbf87.6\%$ & $78.3\%$ & $70.7\%$& $90.9\%$ AttWalk\\
Bar stool &  $74.6\%$ & $\mathbf83.2\%$ & $64.8\%$  & $80.0\%$ MeshWalker  \\
Kids Bed & $70.5\%$ & $68.3\%$ & $\mathbf80.2\%$ & $91.7\%$ MeshNet \\
  \hline
  \end{tabular}
\vspace{0.1in}
\caption{
\textbf{Expert selection at inference.}
Each expert excelled on different classes (3D-FUTURE): AttWalk on Armchair, MeshWalker on Bar stool, and MeshNet~\cite{feng2019meshnet} on Kids Bed. 
During inference the gate selected the best model for each class, e.g. AttWalk's classification for $90.9\%$ of the Armchair.
}
\label{tbl:specific classes}
\end{center}
\end{table}



\begin{table*}[t]
\begin{center}
\small
\begin{tabular}{|l|c|c|c|c|c|c|c|}
\hline
\multirow{2}{*}{Class} & \multicolumn{2}{c|}{AttWalk (Armchair)}  & \multicolumn{2}{c|}{MeshWalker (Bar stool)} & \multicolumn{2}{c|}{MeshNet (Kids Bed)} \\ 
\cline{2-7}
& Before & After & Before & After & Before & After \\
\hline\hline
Armchair & $80.4\%$ & $\mathbf{87.6}\%$ &  $79.4\%$ &  $78.3\%$ & $71.5\%$ &  $70.7\%$ \\

Bar stool & $76.1\%$ & $74.6\%$ & $75.2\%$  & $\mathbf{83.2}\%$ & $66.0\%$ & $64.8\%$ \\ 

Kids Bed &  $71.7\%$ & {$70.5\%$} & $69.3\%$ & $68.3\%$ & $72.7\%$ & $\mathbf{80.2}\%$\\
\hline
  \end{tabular}
  \caption{
\textbf{Individual improvement.} 
Our MoE training process improves each expert’s performance on its designated classes; as expected, the performance of the other experts on those classes correspondingly decreases.
}
\label{tbl:individual experts}
\end{center}
\end{table*}

\begin{table}[tb]
\begin{center}
\begin{tabular}{|l|c|}
\hline
Experts & Accuracy \\ 
\hline
\hline
AttWalk~\cite{Ben_Izhak_2022_WACV}  &  $72.1\%$ \\
$3$ AttWalk & $83.3\%$ \\
MeshWalker~\cite{lahav2020meshwalker}  & $70.2\%$\\
$3$ MeshWalker & $82.9\%$ \\
MeshNet~\cite{feng2019meshnet} & $64.1\%$\\
$3$ MeshNet & $75.6\%$ \\
Heterogeneous (Ours) & $\mathbf{86.1\%}$ \\
 \hline
\end{tabular}
\caption{
\textbf{Comparison of homogeneous and heterogeneous configurations.}
While multiple identical experts can still specialise and improve performance, the heterogeneous setting yields the best results due to architectural diversity.
}
\label{tbl:same experts}
\end{center}
\end{table}

\noindent
\textbf{On the choice of the loss.} 
We aim to evaluate whether adding the similarity loss to the diversity loss leads to better performance. 
Furthermore, we test several  similarity alternative loss functions, beyond KLD. 
Table~\ref{tbl:different similiarity losses} shows that using similarity improves the performance compared to relying on diversity only, with the KLD loss achieving the highest accuracy among the tested similarity losses.

\noindent
\textbf{Expert selection at inference.}
We analyzed the expert selection on 3D-FUTURE when using MeshWalker, AttWalk, and MeshNet as experts. 
Table~\ref{tbl:specific classes} shows the results on three classes.
Each expert excelled on different classes: AttWalk~\cite{Ben_Izhak_2022_WACV} performed best on Armchair, MeshWalker~\cite{lahav2020meshwalker} on Bar stool, and MeshNet~\cite{feng2019meshnet} on Kids Bed. 
As expected, during inference the gate selected the best model for each class:  AttWalk's classification for $90.9\%$ of the Armchair, MeshWalker’s for $80.0\%$ of the Bar stool, and MeshNet’s for $91.7\%$ of the Kids Bed.
We observed that most errors in expert selection occur when the object is atypical. 
For example, a different expert was selected for the few cases of rounded armchairs, whereas most armchairs in the class have straight lines.

\noindent
\textbf{Number of experts.}
We examine the impact of the number of experts on performance.
In Section~\ref{sec:results}, we used three distinct experts, achieving $86.1\%$ accuracy on 3D-FUTURE and  $92.9\%$ on ModelNet40.
Using only two experts (AttWalk and MeshNet) results in slightly lower accuracies: $85.9\%$ and $92.8\%$, respectively.
Adding a fourth expert (WalkFormer~\cite{guo2023walkformer}) yields a marginal gain on 3D-FUTURE ($86.2\%$) and no change on ModelNet40 ($92.9\%$).
These results suggest that the largest improvement comes from combining even two diverse experts, while adding more can still provide incremental gains.

\noindent
\textbf{Improvement of individual experts.}
The gating mechanism not only selects the appropriate expert, but also drives each expert to improve individually throughout the MoE training process and to specialize in specific classes.
Table~\ref{tbl:individual experts} illustrates the performance gains of each expert on its specialized class in the 3D-FUTURE dataset.
As expected, the remaining experts show reduced performance on those same classes.
Consequently, the gating network is selecting among enhanced versions of the original models.

\noindent
\textbf{Homogeneous experts.}
We investigated whether using identical models as experts, rather than heterogeneous ones, could be beneficial.
To evaluate the homogeneous setting, we used three instances of each expert independently.
As shown in Table~\ref{tbl:same experts}, using multiple experts, even of the same type, already leads to improved performance.
This can be attributed to the model's ability to prompt diversity among its identical experts over time.
However, the best results are achieved in the heterogeneous setting, where each expert is inherently suited to different aspects of the data due to its architectural design.

\noindent
\textbf{Pre-training.}
We tested the effectiveness of our pre-training stage, where the gate learns to identify regions important to each expert.
With pre-training, our method achieved $86.1\%$ accuracy, while removing it led to a drop to $83.4\%$.
These results highlight the importance of the pre-training phase for accurate expert selection.

\noindent
\textbf{Running time.} 
When running our system with $3$ experts: MeshWalker, AttWalk, and MeshNet on 3D-Future, 
training using a GTX 1080 Ti GPU takes approximately  $21$ minutes per epoch.
In comparison, on the same GPU, MeshWalker requires $7.5$ minutes, AttWalk $8$ minutes, and MeshNet $7$ minutes per epoch.
The inference time for a given mesh approximately doubles in our system. 
On average, our method takes $270$ milliseconds per mesh, compared to $128$ milliseconds for MeshWalker, $140$ milliseconds for AttWalk, and $109$ milliseconds for MeshNet. 
This increase in inference time is primarily due to the additional computation performed by the gating mechanism prior to selecting the appropriate expert.
However, it is worth noting that, thanks to the pre-trained networks, convergence is typically achieved within just  $10$-$15$ epochs {compared to $90$ epochs for MeshWalker, $97$ for AttWalk and $100$ for MeshNet.}

\noindent
\textbf{Limitations.} 
The primary limitation of our method lies in its increased training and inference time, resulting from the use of multiple models alongside the reinforcement learning agent.
However, the accuracy improves significantly on 3D-FUTURE---by $15.9\%$ over MeshWalker (as reported in~\cite{singh2021meshnet++}), $14.0\%$ over AttWalk (as reported in their paper), and $22.0\%$ over MeshNet (as reported in~\cite{singh2021meshnet++}).

\section{Conclusions}

In this work, we introduce a novel Mixture of Experts (MoE) framework that unifies the strengths of diverse 3D mesh analysis approaches.
At the core of our method is an attention-based gating mechanism that operates directly on random walks over the mesh surface, enabling informed expert selection.

We further propose adding a similarity loss alongside the commonly used diversity loss.
Since balancing these two opposing objectives is inherently challenging, we introduce a reinforcement learning–based training strategy that dynamically adjusts the balance between them.
RL is particularly well-suited for this task, as it is designed to make sequential decisions that optimize long-term outcomes, enabling more effective coordination between diversity and similarity.

Our approach achieves state-of-the-art results across three fundamental mesh analysis tasks: classification, retrieval, and semantic segmentation.
These results are demonstrated on several widely used datasets.

\newpage
\bibliographystyle{eg-alpha-doi} 
\bibliography{main}

\end{document}